\documentclass[10pt,twocolumn,letterpaper]{article}

\usepackage[pagenumbers]{cvpr}              % To produce the CAMERA-READY version
\definecolor{cvprblue}{rgb}{0.21,0.49,0.74}
\usepackage[pagebackref,breaklinks,colorlinks,allcolors=cvprblue]{hyperref}

\def\confName{CVPR}
\def\confYear{2026}

\usepackage{CJKutf8}

\usepackage{hyperref}
\usepackage{url}
\usepackage{enumitem}     
\usepackage{setspace}
\usepackage{graphicx}      % 必需：用于插入图片
\usepackage{float}         % 可选：提供 [H] 等浮动体定位选项
\usepackage{caption}       % 可选：增强标题格式控制
\usepackage{subcaption}    % 可选：用于子图环境（如果有多图）
\usepackage{times}  % DO NOT CHANGE THIS
\usepackage{helvet}  % DO NOT CHANGE THIS
\usepackage{courier}  % DO NOT CHANGE THIS
\usepackage[hyphens]{url}  % DO NOT CHANGE THIS
\usepackage{graphicx} % DO NOT CHANGE THIS
\usepackage{natbib}  % DO NOT CHANGE THIS AND DO NOT ADD ANY OPTIONS TO IT
\usepackage{caption} % DO NOT CHANGE THIS AND DO NOT ADD ANY OPTIONS TO IT
\usepackage{algorithm}
\usepackage{algorithmic}

\usepackage{pifont} %182/183/184
\usepackage{subcaption}
\usepackage{verbatim}
\usepackage{graphicx}
\usepackage{multirow}
\usepackage{xspace}
\usepackage{amsfonts,amssymb} 
\usepackage{microtype}
\usepackage{amsmath}
\usepackage{amsthm}
\usepackage{booktabs}
\usepackage{times}
\usepackage{latexsym}
\usepackage{color}
\usepackage[dvipsnames, table]{xcolor}
\usepackage{soul}
\usepackage{verbatim}
\usepackage{multirow}
\usepackage{xspace}
\usepackage{amsfonts,amssymb} 
\usepackage{microtype}
\usepackage{times}
\usepackage{latexsym}
\usepackage{color}
\usepackage[dvipsnames, table]{xcolor}
\definecolor{lightpurple}{HTML}{CC99FF}
\definecolor{darkgreen}{HTML}{00B050}
\usepackage{soul}
\usepackage{colortbl}
\usepackage{booktabs,makecell, multirow, tabularx}
\usepackage{adjustbox}
\usepackage{xcolor}
\usepackage{listings}
\usepackage{amsmath}

\usepackage{booktabs}
\usepackage{amssymb}
\usepackage{bbding}
\usepackage{pifont}
\usepackage{wasysym}
\usepackage{utfsym}
\usepackage{fontawesome}

\usepackage{mdframed}
\definecolor{arrowgreen}{RGB}{146, 208, 80}

\title{EpiAgent: An Agent-Centric System for Ancient Inscription Restoration}

\author{
    Shipeng Zhu$^{1,2}$, 
    Ang Chen$^{1,2} $, 
    Na Nie$^{4,5}$, 
    Pengfei Fang$^{1,2}$, 
    Min-Ling Zhang$^{1,3}$, 
    Hui Xue$^{1,2}\thanks{ Corresponding author}$ \\
    $^{1}$School of Computer Science and Engineering, Southeast University, China \\
    $^{2}$Key Laboratory of New Generation Artificial Intelligence Technology and \\ Its Interdisciplinary Applications  (Southeast University), Ministry of Education, China \\
    $^{3}$Key Laboratory of Computer Network and Information Integration \\ (Southeast University), Ministry of Education, China \\
    $^{4}$Nanjing University Museum, Nanjing University, China \\
    $^{5}$The China Centre for Linguistic and Strategic Studies, Nanjing University, China \\
    {\tt\small \{shipengzhu,chenang121\}@seu.edu.cn} {\tt\small,} {\tt\small niena@nju.edu.cn} {\tt\small,} {\tt\small\{fangpengfei,zhangml,hxue\}@seu.edu.cn}
}
\begin{document}
\maketitle

% \twocolumn[
% \maketitle
% \begin{center}
%     \includegraphics[width=0.95\linewidth]{sec/img/Fig1.png}
%     \captionof{figure}{Our method restores camera-captured document images in the wild.}
%     \label{fig:teaser}
% \end{center}
% ]

\begin{abstract}

Ancient inscriptions, as repositories of cultural memory, have suffered from centuries of environmental and human-induced degradation. Restoring their intertwined visual and textual integrity poses one of the most demanding challenges in digital heritage preservation. However, existing AI-based approaches often rely on rigid pipelines, struggling to generalize across such complex and heterogeneous real-world degradations.
Inspired by the skill-coordinated workflow of human epigraphers, we propose EpiAgent, an agent-centric system that formulates inscription restoration as a hierarchical planning problem.
Following an Observe–Conceive–Execute–Reevaluate paradigm, an LLM-based central planner orchestrates collaboration among multimodal analysis, historical experience, specialized restoration tools, and iterative self-refinement. This agent-centric coordination enables a flexible and adaptive restoration process beyond conventional single-pass methods.
Across real-world degraded inscriptions, EpiAgent achieves superior restoration quality and stronger generalization compared to existing methods. Our work marks an important step toward expert-level agent-driven restoration of cultural heritage. The code is available at \href{https://github.com/blackprotoss/EpiAgent}{{https://github.com/blackprotoss/EpiAgent}}.

\end{abstract}

\section{Introduction}
\label{sec:intro}

\begin{figure}[!t]
\centering         
\includegraphics[width=1.0\linewidth]{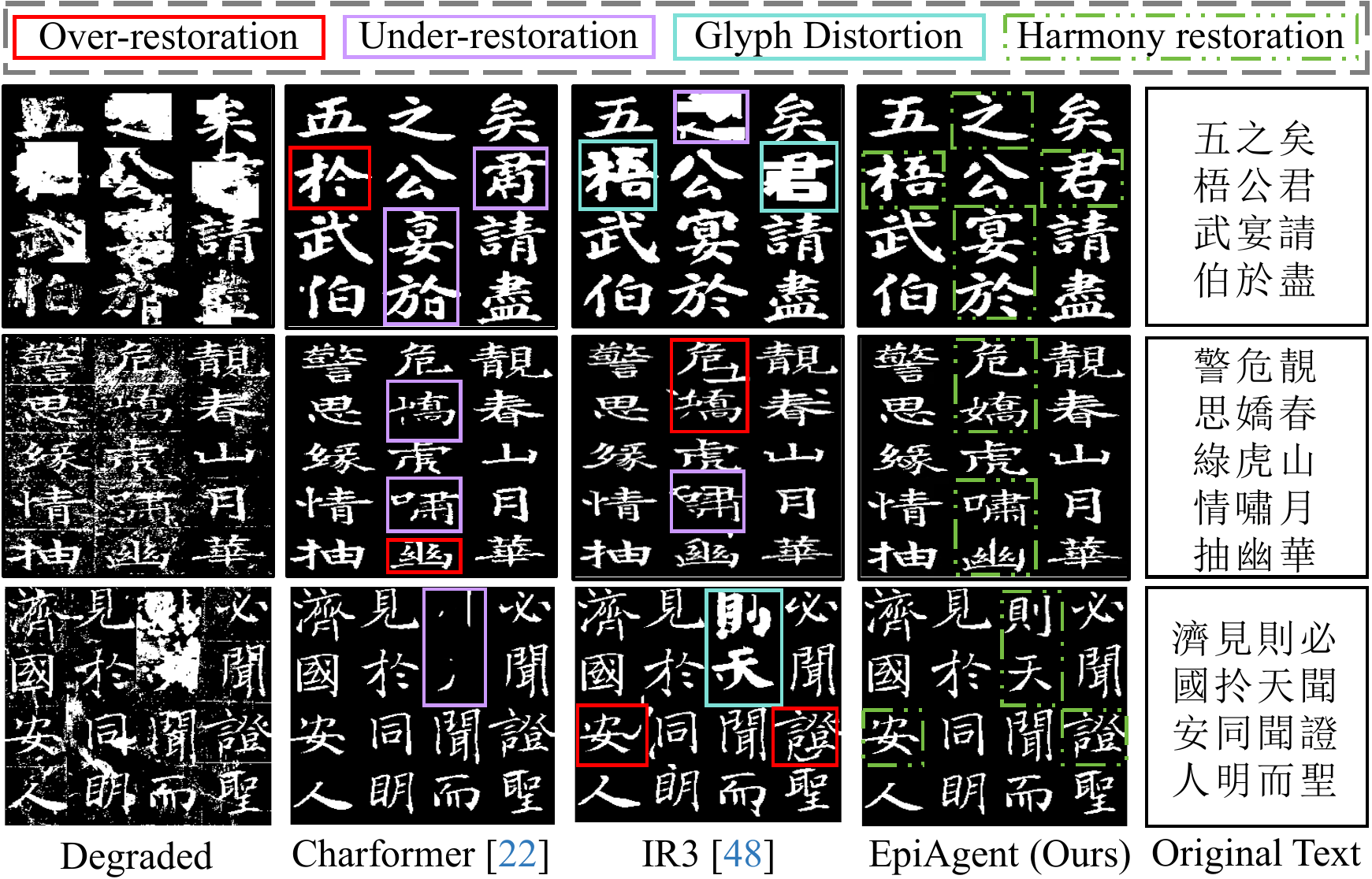}
\caption{Illustration of the restored ancient inscription samples.}
\label{fig:teaser}
\end{figure}

From antiquity to the present, civilizations have inscribed their characters onto diverse materials to preserve information across time and space.
Among these practices, ancient inscriptions, whether carved in stone or preserved as paper rubbings~\cite{starr2018black}, encode an invaluable dual heritage: the irreplaceable textual records and the artistic essence of historical calligraphy from human civilization.
Yet, these cultural artifacts are constantly threatened by environmental decay, material damage, and human intervention. 
The resulting coupled degradations pose a uniquely challenging problem: recovering both semantic integrity and glyph morphology under heterogeneous degradation patterns.
Consequently, inscription restoration represents a critical frontier in digital humanities. Success in this domain would not only recover lost knowledge but also pioneer new paradigms for the intelligent preservation of documentary heritage~\cite{purificato2025role}.

Historically, trained over decades in script typology and artifact conservation, human epigraphers have been central to reconstructing both the textual content and visual form of ancient inscriptions~\cite{matsumoto2022archaeology}. However, the sheer volume of extant materials and the pace of ongoing degradation far exceed expert capacity. In response, recent AI-driven approaches have emerged to automate visual restoration~\cite{sommerschield2023machine}. In particular, most efforts have focused on single-character restoration~\cite{shi2022charformer,zheng2023ea,duan2024restoring}. These approaches do not scale to inscription-level cases, where degradation is spatially coupled and can fully obscure glyph sequences across a tablet.
More recently, researchers have attempted full-inscription restoration using fixed pipelines with predefined workflows~\cite{zhu2024reproducing,zhang2025reviving}. However, this one-size-fits-all strategy lacks adaptability to heterogeneous degradation patterns. 
At a fundamental level, these methods are rooted in the image-to-image transfer paradigm, where the direct mapping from degraded to pristine characters often distorts the original glyph. As shown in Fig.~\ref{fig:teaser}, the resulting over-/under- restorations leave a critical gap between automated techniques and authentic restoration.

\begin{figure}[!t]
\centering         
\includegraphics[width=1.0\linewidth]{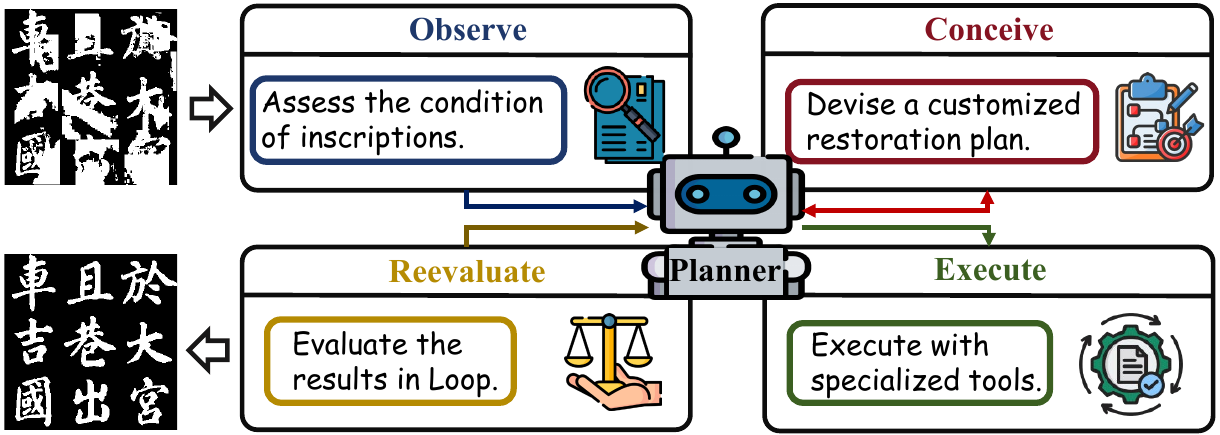}
\caption{Illustration of the EpiAgent framework, which mimics the restoration workflow of human epigraphers.}
\label{fig:archi}
\end{figure}

Therefore, we revisit the complete workflow of human epigraphers, where fine-grained analysis, specialized skills, and aesthetic judgment are orchestrated into a coherent reasoning process. Pursuing agentic systems that learn from expert behavior is thus both natural and necessary for preserving \textit{textual authenticity} and \textit{visual fidelity}. However, existing frameworks largely identify degradations and invoke generic tools~\cite{chen2024restoreagent,zhu2025an}. By contrast, epigraphic restoration demands a hierarchical process along three dimensions:

(1) \textit{Multi-modal Analysis under Complex Degradation.} Inscriptions exhibit spatially varying, structurally entangled, and multi-scale degradations. The restoration system must therefore conduct multi-modal analysis: localizing characters, assessing precise visual damage, deciphering corrupted text, and reconciling results with the historical corpus. These requirements exceed the single degradation-aware schemes used for natural images.
(2) \textit{Adaptive Planning of Specialized Tools.}  Inscription restoration seeks visual–textual harmony rather than isolated enhancement. This calls for a flexible and composable set of specialized tools, which can operate individually or be dynamically composed into task-specific routines.
The system must then weigh evidence from text and appearance to invoke its toolkit, navigating the trade-off between textual authenticity and visual fidelity. Such state-dependent planning is incompatible with fixed pipelines.
(3) \textit{Multi-perspective Evaluation for Self-Refinement.} 
Authentic restoration requires judging not only pixel quality but also textual accuracy and aesthetic consistency, often with third-party expert review.
Incorporating such perspectives into the decision loop enables iterative replanning, progressively steering the system toward expert-aligned epigraphic deliberation. This level of refinement is beyond the reach of single-pass or quality–centric pipelines.

Considering these challenges, we propose \textbf{EpiAgent}, an agent-centric system for ancient inscription restoration. 
At its core lies an LLM-based \textbf{Central Planner} that integrates both generalist and specialist analysis–restoration skills, accumulated restoration experience, and multi-perspective self-reflection.
Operating within an \textit{Observe–Conceive–Execute–Reevaluate} loop, the planner mirrors the collaborative workflow of human epigraphers and drives hierarchical closed-loop decision making throughout the process, as shown in Fig.~\ref{fig:archi}.
In the \textbf{Observe} stage, the planner collects multimodal signals from subordinate generalist–specialist hybrid modules and a historical corpus to establish a structured understanding of the inscription. Then, the \textbf{Conceive} stage fuses these cues with experience distilled from previous executions, thereby devising a customized restoration plan adaptively. 
During \textbf{Execution}, the planner governs a specialized modular toolkit that can be invoked individually or in combination to address complex and coupled degradations.
Finally, in the \textbf{Reevaluate} stage, it closes the loop by evaluating the restored result using automatic metrics and optional expert feedback, updating its plan for subsequent iterations. Extensive experiments demonstrate that EpiAgent effectively handles real-world inscription degradation, achieving notable improvements in both textual authenticity and visual fidelity. 

In a nutshell, the contributions are as follows:

\begin{itemize}

\item We introduce EpiAgent, a pioneering agent-centric system that formalizes the workflow of epigraphers within a unified Observe–Conceive–Execute–Reevaluate paradigm. An LLM-based central planner integrates multi-modal analysis, Specialized tools, and multi-perspective evaluation, enabling hierarchical closed-loop refinement.

\item We decompose inscription restoration into atomic multi-modal operations. This allows our planner to dynamically assemble and schedule a specialized toolkit based on contextual analysis of degradation patterns and historical metadata, addressing complex coupled failures beyond the reach of static pipelines.

\item Extensive experiments and ablation studies demonstrate the superior performance of EpiAgent over existing methods. Our work yields concrete insights for developing expert-level AI systems in cultural heritage preservation.

\end{itemize}

\section{Related Work}
\label{sec:related}

\begin{figure*}[!t]
\centering         
\includegraphics[width=0.9\linewidth]{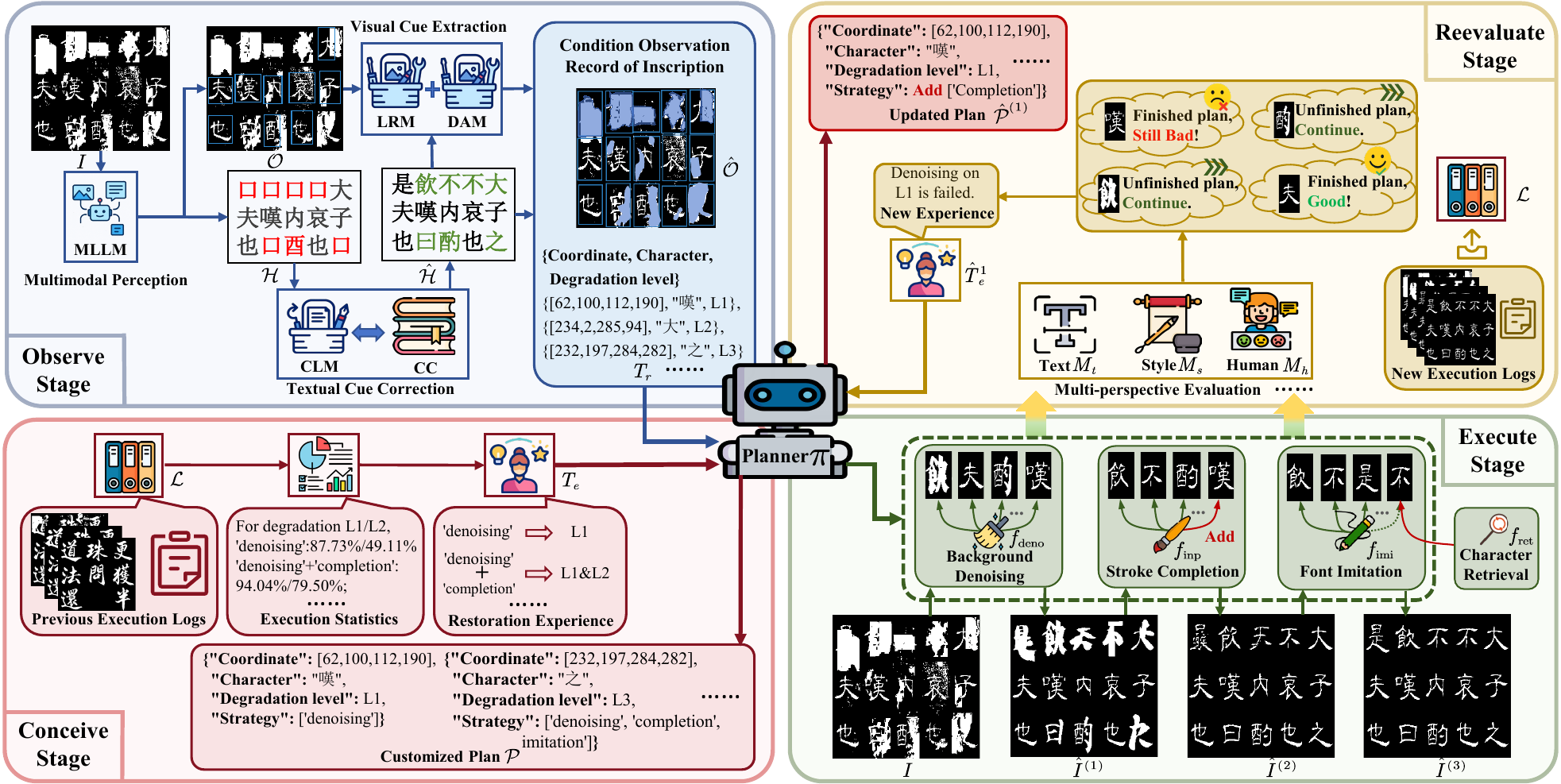}
\caption{Illustration of the workflow of EpiAgent. The ``MLLM", ``LRM", ``DAM", ``CLM", and ``CC" denote Multimodal Large Language Model, Layout Rectification Module, Degradation Assessment Model, Corrective Language Model, and Chinese Corpora, respectively.}
\label{fig:archi}
\end{figure*}
\subsection{Unified Image Restoration}

Modern image restoration has shifted toward unified architectures for coupled degradations in natural images~\cite{jiang2025survey}. Early backbones~\cite{zamir2022restormer,guo2024mambair} enabled unified modeling but still relied on manual degradation identification. Subsequent work moved toward universal restoration, where a single model adapts to diverse patterns. For example, PromptIR~\cite{potlapalli2023promptir} injects degradation-aware prompts, while MoCE-IR~\cite{zamfir2025complexity} routes inputs via mixture-of-experts. With the rise of MLLMs, a new agentic paradigm has emerged. These methods leverage MLLMs to perceive degradation and dynamically invoke pre-trained restoration tools~\cite{chen2024restoreagent,zhu2025an,jiang2025multi,li2025hybrid}.
Despite their flexibility, such general-purpose systems remain constrained in domain-specific contexts~\cite{dong2025phydae}.
In ancient inscription restoration, this limitation becomes fundamental. Wherein, general degradation perception modules cannot capture the linguistic and stylistic nuances of ancient scripts, while off-the-shelf tools cannot preserve calligraphic authenticity. These challenges highlight the need for a domain-tailored framework.

\subsection{Text Image Enhancement}

Text images are inherently multimodal, with their semantics closely tied to visual structure~\cite{wei2025glyphsr}. In real-world scenarios, these images often suffer from multi-causal degradations~\cite{shu2025visual}. Researchers thus explore restoration at multiple granularities. At the character level, CNN/Transformer models reconstruct the fine-grained strokes of individual characters~\cite{wang2021chinese,shi2022charformer}. At the word level, methods typically incorporate structural priors to map degraded words to their clean counterparts~\cite{sun2022tsinit,zhu2024text}. More recently, the focus has expanded to the document level, progressing toward unified restoration frameworks. For instance, DocDiff~\cite{yang2023docdiff} employs frequency-domain priors for multi-task restoration, while DocRes~\cite{zhang2024docres} leverages a Restormer~\cite{zamir2022restormer} to iteratively handle diverse degradations.  However, most existing methods rely on image-to-image translation~\cite{jiang2025survey}, which often compromises structural fidelity and introduces glyph distortion. 
Such drawbacks become unacceptable for inscriptions, where calligraphic consistency is integral to authenticity.

\subsection{Ancient Inscription Restoration}
AI-driven ancient script restoration has become a growing focus in digital humanities~\cite{matsumoto2022archaeology,diao2025oracle,yang2025predicting,cao2022character}. Among these, inscriptions are emerging as key research targets due to their dual nature as historical archives and calligraphic artifacts. 
Early research emphasized text deciphering and attribute classification~\cite{assael2019restoring,assael2022restoring,papavassileiou2023generative}, but neglected visual authenticity. This omission is a critical flaw for ideographic scripts, where semantics and morphology are inseparable~\cite{zhu2024reproducing}. Subsequent visual restoration approaches focused on character-level enhancement~\cite{zheng2023ea,liu2025structural}, yet failed to maintain inscription-level coherence. Recent advances have extended to full-scale restoration: Duan et al.~\cite{duan2024restoring} introduced a context-aware approach limited to short phrases. Zhu et al.~\cite{zhu2024reproducing} proposed a global-local framework for full-inscription restoration that suffers from error propagation. 
In parallel, AutoHDR~\cite{zhang2025reviving} has shown potential for damaged content prediction using LLMs, though its style-transfer backbones may cause calligraphic distortion under heterogeneous degradation.

\begin{figure*}[!t]
\centering         
\includegraphics[width=0.9\linewidth]{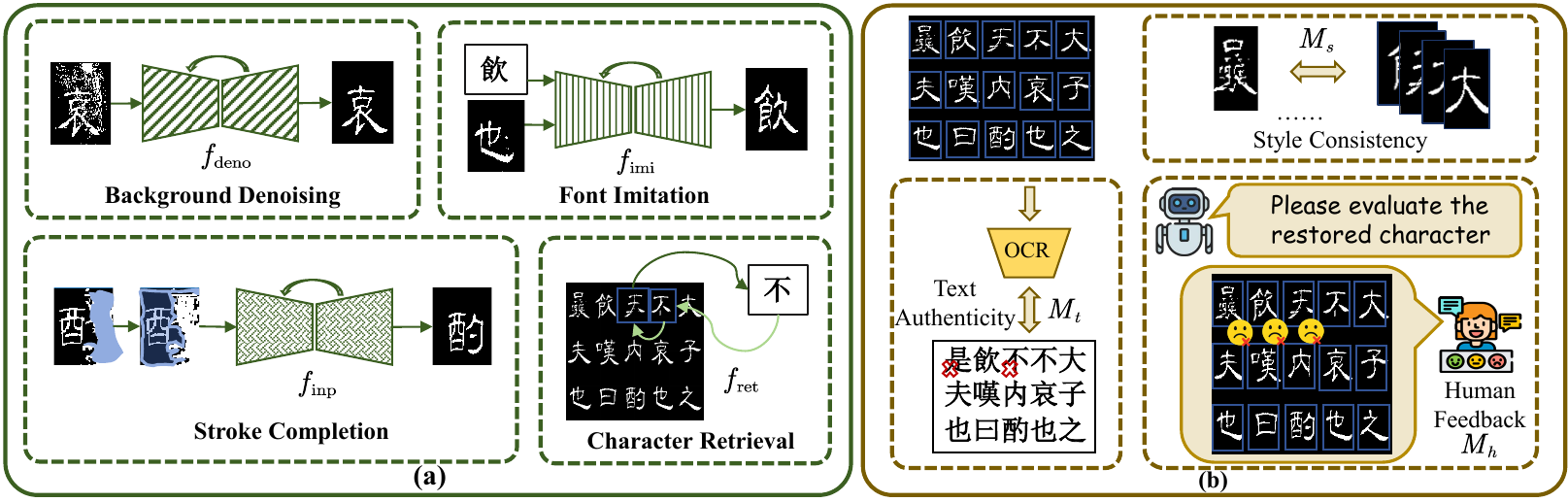}
\caption{(a) Process of Specialized Restoration Tools; (b) Details of Multi-perspective Evaluation.}
\label{fig:module}
\end{figure*}

\section{EpiAgent}
\label{sec:method}

As illustrated in Fig.~\ref{fig:archi}, EpiAgent is an \textit{agent-centric restoration system} that operationalizes the deliberative workflow of human epigraphers. Given a degraded inscription image $I$ affected by coupled degradation factors $\mathcal{D}$, our objective is to produce a restored output $\hat{I}$ that maximizes both textual authenticity and visual fidelity. Formally, the restoration process is governed by a \textbf{central planner} $\pi$, implemented as an agentic LLM, Kimi-K2~\cite{team2025kimi}, which dynamically orchestrates analysis, reasoning, and tool invocation. Departing from one-pass pipelines, EpiAgent follows a dynamic four-stage process, \textit{Observe-Conceive-Execute-Reevaluate}, which is executed iteratively until a stopping criterion.

Let $\mathcal{T}$ denote the restoration trajectory, comprising condition assessment $T_r$, experience priors $T_e$, execution plan $P$, and evaluation metrics $\mathcal{M}$. The central planner $\pi$ operates on $\mathcal{T}$ to generate action sequences $\mathcal{F}$ adaptively. This closed-loop orchestration enables EpiAgent to evolve from reactive execution toward expert-informed deliberation by coupling procedural reasoning with domain-specific expertise.

\subsection{Observe Stage}

An effective restoration plan requires a fine-grained assessment of the textual content, calligraphic style, and degradation patterns in inscription $I$. Accordingly, the Observe stage builds a comprehensive record \(T_r\) via a two-step scheme:

\noindent\textbf{Step 1: General multimodal perception.}
An MLLM~\cite{guo2025seed1} produces an initial layout $\mathcal{O}$ and textual hypotheses \(\mathcal{H}\).

\noindent\textbf{Step 2: Specialized visual and textual refinement.}
\textbf{(a) Text Cue Correction.} A Corrective Language Model (CLM), fine-tuned~\cite{hu2022lora} on a 7B LLM~\cite{team2024qwen2} and equipped with Retrieval-Augmented Generation (RAG)~\cite{xiao2023bge}, queries a large-scale Chinese corpus to produce the corrected reading \(\hat{\mathcal{H}}\). Notably, the system allows human experts to verify $\hat{\mathcal{L}}$ for authentic correction.
\textbf{(b) Visual Cue Extraction.} First, a Layout Rectification Module (LRM) consumes \(\mathcal{O}\) and \(\hat{\mathcal{H}}\) to predict a rectified layout $\hat{\mathcal{O}}$ that can explicitly account for fully missing or occluded regions. Second, a Degradation Assessment Module (DAM) delineates pixel-level degradation segmentation masks \(\mathcal{S}_d\) and assigns discrete severity levels (none, slight, middle, severe) as \(s \in \{0,1,2,3\}\). 
Therefore, we formalize the observation record as: $T_r = \langle I, \mathcal{S}_d, s, \hat{\mathcal{H}}, \hat{\mathcal{O}} \rangle.$

\subsection{Conceive Stage}
Having observed the inscription status, the agent must translate this assessment \(T_r\) into an actionable plan \(\mathcal{P}\). That is, it must decide how to select and sequence restoration tools for each character \(c \in \mathcal{C}\), where \(\mathcal{C}\) denotes the set of visual characters in the corrected reading \(\hat{\mathcal{H}}\). Notably, EpiAgent plans at the level of fine-grained character restoration. Reliable layout and text predictions from the \textbf{Observe} stage isolate character regions, allowing global removal of cross-character background noise and leaving only local residuals for specialized character-level tools.
Rather than relying on trial-and-error, the planner \(\pi\) should exploit previous restoration experience \(T_e\) to make adaptive decisions.

Concretely, \(T_e\) is distilled from historical execution logs \(\mathcal{L}\). We mine \(\mathcal{L}\) to extract statistical priors that map degradation patterns \(\mathcal{S}_d\) to tool-efficacy distributions \(p(f \mid \mathcal{S}_d)\), where \(f \in \mathcal{F}\) denotes a restoration tool.
For each character \(c \in \mathcal{C}\), the planner \(\pi\) conditions on the concatenated input \([T_r; T_e]\) and produces an individual action sequence:
\begin{equation}
P_c \;=\; \pi(T_r, T_e, c)
      \;=\; \bigl(f_1^{(c)}, f_2^{(c)},..., f_{N_c}^{(c)} \bigr),
\end{equation}
where each \(f_i^{(c)}\) is a selected restoration tool for \(c\), and \(N_c\) denotes the character-dependent length of the sequence; the index \(i\) specifies the execution order for \(c\). The overall plan for inscription \(I\) is then given by $\mathcal{P} \;=\; \{\, P_c \,\}_{c \in \mathcal{C}}.$

% %%%%%%%%%%%%%%%%%%%颜色定义
\definecolor{color_gray}{RGB}{229,229,229}
\definecolor{color_blue}{RGB}{252,182,165}
\definecolor{color_pink}{RGB}{255,217,178}
\definecolor{color_yellow}{RGB}{255,255,204}
\definecolor{color_blue1}{RGB}{135, 206, 235}
\cellcolor{color_blue} \cellcolor{color_pink} \cellcolor{color_yellow} \cellcolor{blue!8}
% %%%%%%%%%%%%%%%%%%%

\begin{table*}[ht!]
\centering
\caption{Inscription image restoration results on Testing Set S, R-I, and R-II. Comparison with state-of-the-art methods. The best and the second-best results are \textbf{highlighted} and \underline{underlined}.}
\setlength{\tabcolsep}{2.8mm} %2.2mm
\renewcommand{\arraystretch}{1}
\adjustbox{width=1\linewidth}{
   \begin{tabular}{lcccc ccc cccc}
       \toprule
            \multicolumn{12}{c}{\textbf{(a) Inscription Image Restoration on Testing Set S}}\\
            \cmidrule(lr){1-12}
%%%%%%%%%%                 
            \multirow[c]{2}{*}{\vspace*{-1.5mm}\centering Method / Metric} & \multicolumn{7}{c}{Quality} & \multicolumn{3}{c}{Recognition} & \multicolumn{1}{c}{End-to-End} \\
            \cmidrule(lr){2-8} \cmidrule(lr){9-11} \cmidrule(lr){12-12}
            % \cline{2-8} \cline{9-11} \cline{12-12}
            % \cmidrule(lr){2-8} \cmidrule(lr){9-11} \cmidrule(lr){12-12}
                                     & PSNR $\uparrow$ &SSIM $\uparrow$ &LPIPS $\downarrow$  &CLIP-IQA $\uparrow$  &MUSIQ $\uparrow$  & MANIQA $\uparrow$ &NIMA $\uparrow$  &Top-1 Acc. $\uparrow$   &Top-5 Acc. $\uparrow$  &Macro Acc. $\uparrow$  &1-NED $\uparrow$ \\
%%%%%%%%%%                 
\midrule[0.5pt]  
CharFormer~\cite{shi2022charformer}    &19.74   &0.9503 &0.0478 &0.8763 &52.77 &0.4339 &0.5547 & 0.9109  & 0.9533 & 0.5549  & 0.8313 \\
DocDiff~\cite{yang2023docdiff}       &20.61   &0.9565 &\underline{0.0361} &0.8962 &53.31 &0.4444 &\underline{0.5559}  & 0.9275  & 0.9622 & 0.5697 & 0.8439  \\
GSDM~\cite{zhu2024text}          &20.37   &0.9495 &0.0390 &0.8933 &53.13 &0.4422 &0.5550  & 0.8948  & 0.9414  & 0.5368  & 0.8093 \\
% SeedEdit3.0~\cite{wang2025seededit}   &15.18   &0.8414 &0.2908 &0.8529 &51.12 &0.4012 &0.5046  & 0.7732  & 0.8754  & 0.4292  & 0.3870  \\ 
Restormer~\cite{zamir2022restormer}     &18.90   &0.9390 &0.0667 &0.8891 &52.82 &0.4391 &0.5509  & 0.8097  & 0.8824  & 0.4454  & 0.7523 \\
MambaIR~\cite{guo2024mambair}       &21.10   &\underline{0.9599} &0.0377 &0.8923 &53.22 &\underline{0.4446} &0.5556   & 0.9093  & 0.9492  & 0.5561  & 0.8251  \\
PromptIR~\cite{potlapalli2023promptir}      &19.30   &0.9464 &0.0551 &0.8882 &52.95 &0.4390 &0.5542   & 0.8601  & 0.9214  & 0.4970  & 0.7741  \\
MoCE-IR~\cite{zamfir2025complexity}   &19.39   &0.9462 &0.0473 &0.8955 &53.24 &0.4399 &0.5553   & 0.8147  & 0.8929  & 0.4562  & 0.7526  \\
IR3~\cite{zhu2024reproducing}          &\underline{21.15}   &0.9540 &0.0388 &\underline{0.8987} &\underline{53.35} &0.4429 &0.5547   & \underline{0.9626}  &  \underline{0.9846} & \underline{0.6459}  & \underline{0.8855}  \\
\cellcolor{blue!8}EpiAgent (Ours)      &\cellcolor{blue!8}\textbf{22.14}   &\cellcolor{blue!8}\textbf{0.9684} &\cellcolor{blue!8}\textbf{0.0254} &\cellcolor{blue!8}\textbf{0.9004} &\cellcolor{blue!8}\textbf{53.98} &\cellcolor{blue!8}\textbf{0.4553} &\cellcolor{blue!8}\textbf{0.5576}   & \cellcolor{blue!8}\textbf{0.9889}  & \cellcolor{blue!8}\textbf{0.9942}  & \cellcolor{blue!8}\textbf{0.6877}  & \cellcolor{blue!8}\textbf{0.9069} \\
Intact   & -  & - & - & - & - & - & -  &  0.9971  & 0.9996   & 0.7064   & 0.9120  \\ 
\toprule
            \multicolumn{6}{c}{\textbf{(b) Inscription Image Restoration on Testing Set R-I}}
            &\multicolumn{6}{c}{\textbf{(c) Inscription Image Restoration on Testing Set R-II}}\\
%%%%%%%%%%        
            \cmidrule(lr){1-6} \cmidrule(lr){7-12}
            \multirow[c]{2}{*}{\vspace*{-1.5mm}\centering Method  / Metric} & \multicolumn{4}{c}{Quality} & \multicolumn{1}{c}{End-to-End}  & \multirow[c]{2}{*}{\vspace*{-1.5mm}\centering Method  / Metric} & \multicolumn{4}{c}{Quality} & \multicolumn{1}{c}{End-to-End} \\
            \cmidrule(lr){2-5} \cmidrule(lr){6-6} \cmidrule(lr){8-11} \cmidrule(lr){12-12}
                &CLIP-IQA $\uparrow$ &MUSIQ $\uparrow$ &MANIQA $\uparrow$  &NIMA $\uparrow$  & 1-NED $\uparrow$ &  &CLIP-IQA $\uparrow$ &MUSIQ $\uparrow$ &MANIQA $\uparrow$  &NIMA $\uparrow$ & 1-NED $\uparrow$ \\
%%%%%%%%%%       

\midrule[0.5pt]  
CharFormer~\cite{shi2022charformer} &0.9320 &49.34 &0.3993 &0.5398 &0.5177                          & CharFormer~\cite{shi2022charformer} &0.9277 &49.14 &0.4009 &0.5289 & 0.4642 \\
DocDiff~\cite{yang2023docdiff}    &\underline{0.9375} &49.58 &0.4059 &\underline{0.5408} & 0.5080 & DocDiff~\cite{yang2023docdiff} &0.9352 &49.45 &0.4084 &\underline{0.5307} & 0.4619 \\
GSDM~\cite{zhu2024text}       &0.9330 &49.43 &0.4027 &0.5365 & 0.5245                         & GSDM~\cite{zhu2024text} &\underline{0.9373} &49.26 &0.4042 &0.5302 & 0.4789 \\
% SeedEdit3.0 &0.9111 &47.81 &0.3817 &0.4921 &0.2758                         & SeedEdit3.0 &0.9136 &47.25 &0.3803 &0.4725 & 0.2280  \\
Restormer~\cite{zamir2022restormer}  &0.9256 &49.02 &0.4029 &0.5322 & 0.4866                         & Restormer~\cite{zamir2022restormer} &0.9081 &48.32 &0.4061 &0.5225 & 0.4307 \\
MambaIR~\cite{guo2024mambair}    &0.9346 &49.40 &0.4046 &0.5356 & 0.5205                         & MambaIR~\cite{guo2024mambair} &0.9259 &48.94 &\underline{0.4104} &0.5272 & 0.4697 \\
PromptIR~\cite{potlapalli2023promptir}   &0.9237 &48.31 &0.3951 &0.5273 & 0.4312                         & PromptIR~\cite{potlapalli2023promptir} &0.9030 &47.61 &0.3968 &0.5208 & 0.3968 \\
MoCE-IR~\cite{zamfir2025complexity}     &0.9317 &48.53 &0.3999 &0.5323 & 0.4260                         & MoCE-IR~\cite{zamfir2025complexity}  &0.9185 &48.02 &0.4012 &0.5282 & 0.3892 \\
IR3~\cite{zhu2024reproducing}       &0.9370 &\underline{49.77} &\underline{0.4090} &0.5334 & \underline{0.5539}                  &IR3~\cite{zhu2024reproducing} &0.9346 &\underline{49.58} &0.4082 &0.5298 & \underline{0.5374} \\
\cellcolor{blue!8}EpiAgent (Ours)   &\cellcolor{blue!8}\textbf{0.9393} &\cellcolor{blue!8}\textbf{50.29} &\cellcolor{blue!8}\textbf{0.4179} &\cellcolor{blue!8}\textbf{0.5414} & \cellcolor{blue!8}\textbf{0.5766}               & \cellcolor{blue!8}EpiAgent (Ours) &\cellcolor{blue!8}\textbf{0.9388}   &\cellcolor{blue!8}\textbf{49.94}   &\cellcolor{blue!8}\textbf{0.4157}   &\cellcolor{blue!8}\textbf{0.5381} & \cellcolor{blue!8}\textbf{0.5546} \\
             \bottomrule
       \end{tabular}
   }
   
\label{table:main_compare}
\end{table*}

\subsection{Execute Stage}
Given the complexity of coupled degradations, a single monolithic restorer is prone to under- or over-restoration or glyph distortion. Following the practice of human epigraphers, we instead factor restoration into four specialized composable tools that can be invoked independently or assembled on demand. Specifically, the Execute stage instantiates plan $\mathcal{P}$ via a composable toolkit $\mathcal{F}$. As shown in Fig.~\ref{fig:module}(a), $\mathcal{F}$ comprises three diffusion-based tools~\cite{ho2020denoising}:

\begin{itemize}
\item \textbf{Background Denoising} ($f_{\text{den}}$): Removes surface noise while preserving stroke structures via masked diffusion conditioned on $\mathcal{S}_d$.
\item \textbf{Stroke Completion} ($f_{\text{inp}}$): Performs targeted inpainting of missing or severely degraded regions indicated by $\mathcal{S}_d$, avoiding deformation of intact strokes.
\item \textbf{Font Imitation} ($f_{\text{imi}}$): For heavily corrupted characters, synthesizes stylistically consistent glyphs by learning style priors from high-quality exemplars of the same stele, thus maintaining calligraphic harmony.
\end{itemize}

Additionally, a \textbf{Character Retrieval} module ($f_{\text{ret}}$) serves as a fallback, searching for identical characters within $I$ to replace irreparable ones without introducing style drift. 

Given the per-character sequences \(P_c = \bigl(f_i^{(c)}\bigr)_{i=1}^{N_c}\), the execution process applies the scheduled operators in order. At iteration \(k\), the restored inscription \(\hat{I}^{(k)}\) is obtained by:
\begin{equation}
\hat{I}^{(k)}[c] \;=\; f_{N_c}^{(c)} \circ \cdots \circ f_1^{(c)}\bigl(\hat{I}^{(k-1)}[c]\bigr), \quad \forall\, c \in \mathcal{C},
\end{equation}
where \(\hat{I}^{(0)} = I\). The overall execution trajectory is thus governed by the sequences \(\{P_c\}_{c \in \mathcal{C}}\) and the toolkit \(\mathcal{F}\).

\subsection{Reevaluate Stage}
To ensure both textual and visual harmony, it is crucial to introduce multi-perspective metrics after each execution pass. Such signals provide principled stopping and rollback criteria. Therefore, they allow the central planner to achieve self-refinement by revising plans for under-restored characters and continuously distilling experience for future decisions.
As shown in Fig.~\ref{fig:module}(b), at iteration \(k\) we evaluate each character \(c \in \mathcal{C}\) on the current \(\hat{I}^{(k)}\) using metrics:
\begin{itemize}
\item \textbf{Text Authenticity}. This metric quantifies semantic correctness by comparing OCR text to the corrected reading: $M_t^{(k)}(c) = 1 - \mathrm{CER}\bigl(\mathrm{OCR}(\hat{I}^{(k)}[c]), \hat{\mathcal{H}}[c]\bigr) \in [0,1]$.
\item \textbf{Style Consistency}.This aesthetic metric measures calligraphic conformity to the reference style distribution: $M_s^{(k)}(c) = \mathrm{CosSim}\bigl(\phi(\hat{I}^{(k)}[c]), \phi_{\text{ref}}\bigr) \in [0,1]$, where $\phi_{\text{ref}}$ is a style embedding from high-quality exemplars.
\item \textbf{Human Feedback (Optional)}: This metric provides an expert acceptance signal for ambiguous or high-stakes cases, serving as a hard decision criterion and calibrating thresholds for replanning or termination: $M_h^{(k)}(c) \in \{0,1\}$.
\end{itemize}

Given thresholds \(\tau_t, \tau_s \in (0,1]\) and a maximum iteration budget \(K_{\max}\), the failure set at iteration \(k\) is defined as:
\begin{equation}
\mathcal{F}^{(k)} = \Bigl\{ c \in \mathcal{C} \,\Bigm|\,
\begin{aligned}
& (M_h^{(k)}(c) = 0) \;\lor \\
& (M_t^{(k)}(c) < \tau_t) \;\lor\; (M_s^{(k)}(c) < \tau_s)
\end{aligned}
\Bigr\}.
\end{equation}

If \(\mathcal{F}^{(k)} \neq \varnothing\) and \(k < K_{\max}\), the planner \(\pi\) uses \(\mathcal{F}^{(k)}\), together with \(T_r\) and \(T_e\), to generate a revised plan \(\mathcal{P}^{(k+1)}\) that focuses on the failed characters; otherwise, the process terminates. Finally, we update the execution logs \(\mathcal{L}\) to refine the empirical priors \(T_e\). Thus, the strategy is dynamically updated during inference rather than fixed by a static prior.

\section{Experiments}
\label{sec:experiment}

\begin{figure*}[!t]
\centering         
\includegraphics[width=17cm]{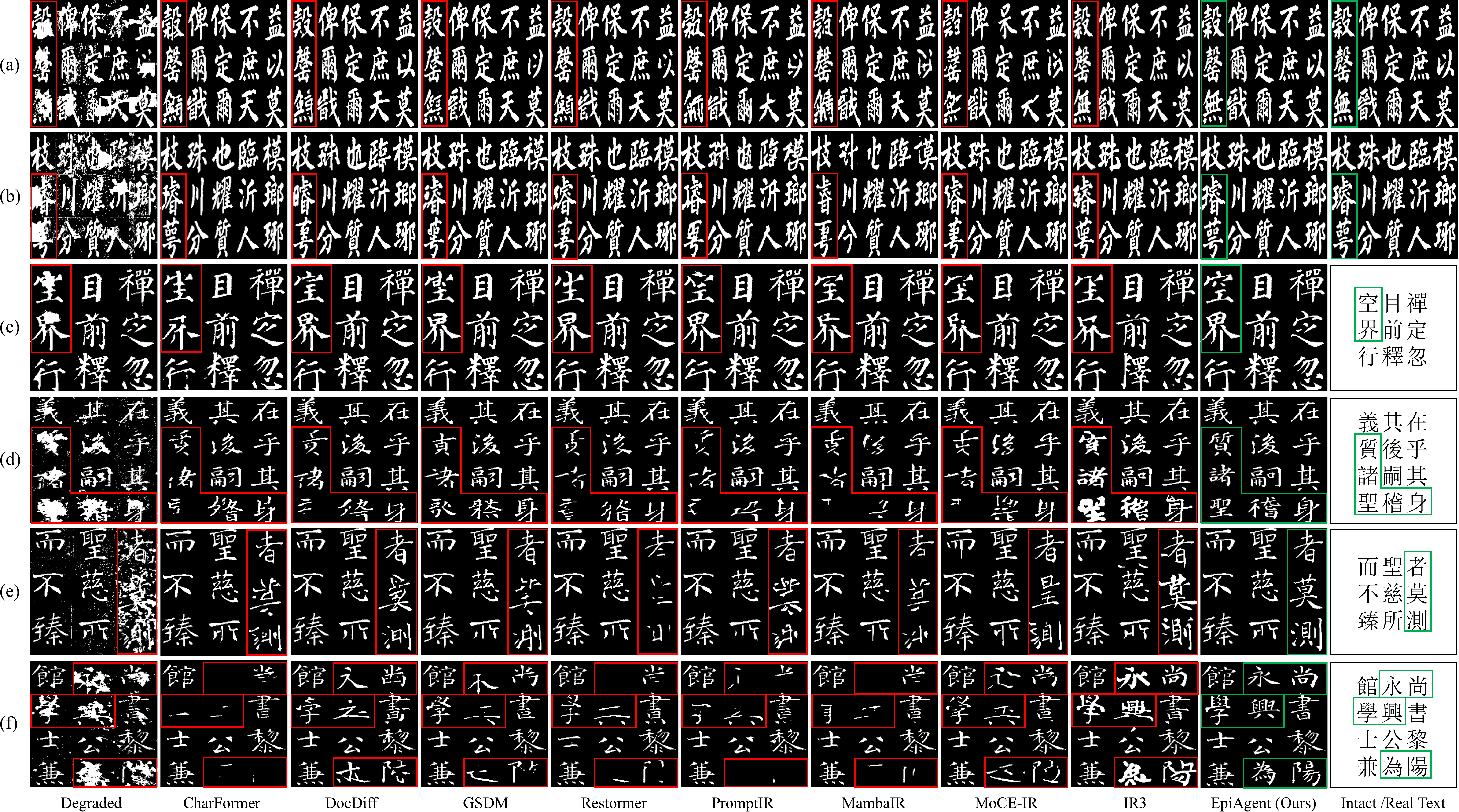}
\caption{Restoration results of different methods on degraded inscription images. (a)-(b) are from Testing Set S, (c)-(d) belong to Testing Set R-I, and (e)-(f) belong to Testing Set R-II. The red borders denote the degraded patches and the restored patches by competing methods, while the green counterparts denote the ground-truth text and the restored patches by EpiAgent.}
\label{fig:main_compare}
\end{figure*}

\subsection{Evaluation Protocol}

We evaluate EpiAgent on the Chinese Inscription Rubbing Images (CIRI) dataset~\cite{zhu2024reproducing}, which comprises a wide range of real inscription rubbings featuring diverse calligraphic styles, complex character structures, and compound degradation patterns. CIRI contains 24k synthetic inscription images (20K for training, 4K for testing) and 2k real rubbings split into two test sets. Type I (R-I) includes images whose degradation patterns are partially reused as sources for synthesizing defects in the synthetic subset, while Type~II (R-II) consists of fragments with entirely unseen degradation that remain unseen during synthesis and training. 
For comprehensive assessment, we adopt seven image quality metrics to evaluate the visual appearance of reconstructed characters: three full-reference metrics (PSNR, SSIM~\cite{wang2004image}, LPIPS~\cite{zhang2018unreasonable}) and four no-reference metrics (CLIP-IQA~\cite{wang2023exploring}, MUSIQ~\cite{ke2021musiq}, MANIQA~\cite{yang2022maniqa}, NIMA~\cite{talebi2018nima}), with particular emphasis on image aesthetics and stylistic fidelity. To quantify glyph fidelity and textual authenticity, we report Top-1 Accuracy, Top-5 Accuracy, and Macro Accuracy~\cite{zhang2023deep} for character-level recognition, as well as 1-NED~\cite{zhang2019icdar} for image-level text similarity between predicted and ground-truth transcriptions.

\subsection{Comparison with State-of-the-Art Methods}

We compare EpiAgent against a series of state-of-the-art open-source methods from related areas. These include Restormer~\cite{zamir2022restormer}, MambaIR~\cite{guo2024mambair}, PromptIR~\cite{potlapalli2023promptir}, and MoCE-IR~\cite{zamfir2025complexity} from unified image restoration; CharFormer~\cite{shi2022charformer}, GSDM~\cite{zhu2024text}, and DocDiff~\cite{yang2023docdiff} from text image enhancement; a tailored baseline~\cite{zhu2024reproducing} (denoted as IR3) from inscription restoration.
All comparison methods are trained on the CIRI training set using their official implementations and default hyperparameters, and their performance is evaluated on the three CIRI test splits. It is worth noting that existing agentic frameworks~\cite{chen2024restoreagent,zhu2025an} for natural images cannot be directly compared, since their general perception modules and restoration tools cannot be readily transferred to inscription restoration without substantial redesign. In addition, recent tailored methods~\cite{yang2025predicting,zhang2025reviving} are difficult to reproduce faithfully due to the unavailability of some components.

\begin{table}[t]
\centering
   \caption{User study of inscription image restoration results. The best and the second best results are \textbf{highlighted} and \underline{underlined}.}
\setlength{\tabcolsep}{2.8mm} %2.2mm
\renewcommand{\arraystretch}{1}
\adjustbox{width=1\linewidth}{
   \begin{tabular}{cccc}
       \toprule

                 \multirow[c]{2}{*}{\vspace*{-1.5mm}\centering Method / Metric} & \multicolumn{3}{c}{User Study (\%) $\uparrow$ } \\
                 \cmidrule[0.5pt](lr){2-4}
                 & Top-1 Ranking  & Top-3 Ranking  & Mean Ranking~\cite{ostertagova2014methodology} \\
        
%%%%%%%%%%%%%% 
\midrule[0.5pt]      
Charformer~\cite{shi2022charformer}      & 3.72 & 26.57 & 52.39 \\
DocDiff~\cite{yang2023docdiff}         & 8.38 & 46.63 & 63.46 \\
GSDM~\cite{zhu2024text}            & 5.75 & 39.55 & 60.70 \\
Restormer~\cite{zamir2022restormer}       & 0.89 & 9.20  & 28.10 \\
MambaIR~\cite{guo2024mambair}         & 3.88 & 32.74 & 57.55 \\
PromptIR~\cite{potlapalli2023promptir}        & 0.63 & 7.75  & 20.96 \\
MoCE-IR~\cite{zamfir2025complexity}         & 1.52 & 11.33 & 36.62 \\
IR3~\cite{zhu2024reproducing}            & \underline{15.60} & \underline{51.14} & \underline{67.41} \\
\cellcolor{blue!8}EpiAgent (Ours)    &\cellcolor{blue!8}\textbf{59.66} &\cellcolor{blue!8}\textbf{84.18} &\cellcolor{blue!8}\textbf{82.11} \\

 \bottomrule
       \end{tabular}
   }
\label{table:user_study_new}
\end{table}

\begin{figure*}[t]
\centering         
\includegraphics[width=0.9\linewidth]{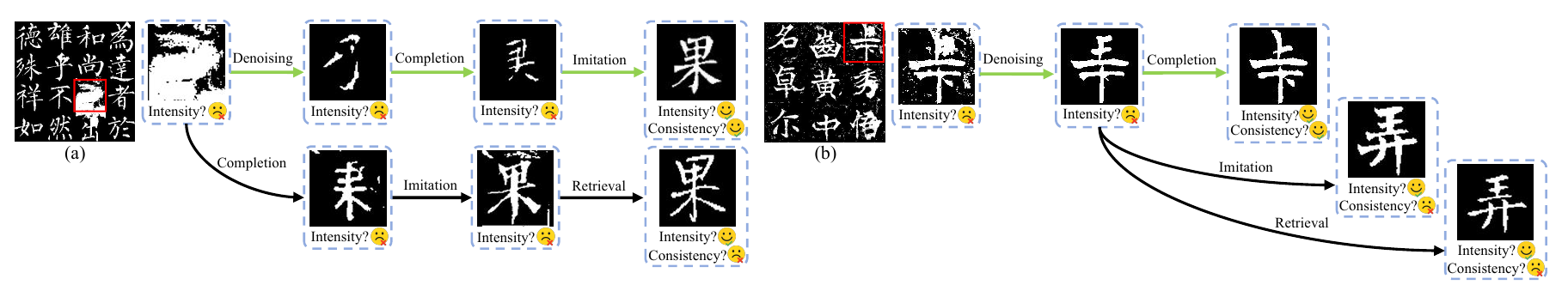}
\caption{Exemplary comparison between different tool invocation sequences faced with (a) severely degraded (L3) and (b) slightly degraded (L1) character blocks. The \textcolor{arrowgreen}{green} lines mark the optimal restoring sequence.}
\label{fig:sequence}
\end{figure*}

We conduct quantitative experiments on the three CIRI test sets to evaluate both the visual fidelity and textual authenticity of images restored by different methods, with results summarized in Tab.~\ref{table:main_compare} and Fig.~\ref{fig:main_compare}. In parallel, we perform a user study involving over twenty epigraphers and scholars who provided professional subjective assessments of restoration quality, as reported in Tab.~\ref{table:user_study_new}.

The results in Tab.~\ref{table:main_compare} show that EpiAgent achieves consistently superior performance on inscription restoration. Across all metrics on the synthetic split S and the real splits R-I and R-II, EpiAgent surpasses all competing methods, demonstrating the advantages of an agent-centric restoration strategy in both recovering the visual appearance of damaged regions and preserving textual authenticity. The user study in Tab.~\ref{table:user_study_new} further demonstrates that human experts overwhelmingly prefer EpiAgent over all baselines.
Together, these results indicate that EpiAgent delivers SOTA performance under both metric- and expert-centered evaluation.

Qualitative comparisons in Fig.~\ref{fig:main_compare} provide additional insights: (1) EpiAgent maintains high restoration quality under both minor degradation (e.g., the region highlighted in Fig.~\ref{fig:main_compare}(c)) and severe degradation (e.g., Fig.~\ref{fig:main_compare}(d)). This robustness arises from degradation-aware and experience-guided planning and execution, which enable flexible deployment of appropriate tools for specific degradation patterns; (2) EpiAgent achieves strong style perception and alignment, both when completing lightly damaged characters (Fig.~\ref{fig:main_compare}(c)) and when fully reconstructing missing characters in large degraded areas (Fig.~\ref{fig:main_compare}(f)). This preservation of calligraphic consistency is crucial for aesthetically convincing inscription restoration; 
(3) EpiAgent demonstrates robust generalization across synthetic and real inscription images. While methods such as CharFormer~\cite{shi2022charformer} and IR3~\cite{zhu2024reproducing} perform competitively on synthetic test images (Fig.~\ref{fig:main_compare}(a),(b)), their performance drops markedly on real inscriptions.

\begin{table}[t]
\centering
   \caption{Ablation studies of the analysis modules used in the Observation stage. The best and the second-best results are \textbf{highlighted} and \underline{underlined}.}
\setlength{\tabcolsep}{2.8mm} %2.2mm
\renewcommand{\arraystretch}{1}
% \adjustbox{width=1\linewidth}{
\resizebox{!}{1.1cm}{
   \begin{tabular}{ccc ccc}
       \toprule

                 \multicolumn{3}{c}{Analysis Module}
                 & \multicolumn{3}{c}{1-NED $\uparrow$}
                  \\ \hline
                \rule{0pt}{2.6ex}
                 MLLM &CLM &RAG &Set S & Set R-I &Set R-II \\
                 \cmidrule[0.5pt](lr){1-3}\cmidrule[0.5pt](lr){4-6}
%%%%%%%%%%%%%% 
% \midrule[0.5pt]      
\ding{51} & \ding{55} & \ding{55}              &0.6481 &0.5509 &0.4336 \\
\ding{51} &\ding{51} & \ding{55}       &\underline{0.9211} &\underline{0.8759} &\underline{0.8486} \\
\cellcolor{blue!8}\ding{51} &\cellcolor{blue!8}\ding{51} &\cellcolor{blue!8}\ding{51}     &\cellcolor{blue!8}\textbf{0.9742} &\cellcolor{blue!8}\textbf{0.9694} &\cellcolor{blue!8}\textbf{0.9606} \\
 \bottomrule
       \end{tabular}
   }

\label{table:observation_new}
\end{table}

\subsection{Ablation Study}

\subsubsection{The Impact of Multimodal Analysis}

Multimodal analysis in the \textbf{Observe} stage is critical to EpiAgent, as it provides a comprehensive assessment of degraded inscriptions on which all downstream planning depends. We therefore ablate the multimodal analysis modules and study their impact on overall restoration performance. The evaluated modules include: (i) an MLLM for general perception, (ii) a fine-tuned Corrective Language Model (CLM) for text correction, and (iii) a Retrieval-Augmented Generation (RAG) module over a Chinese corpus. Tab.~\ref{table:observation_new} reports end-to-end spotting performance under different combinations.
The results reveal three main observations: (1) a standalone MLLM struggles to handle the complex visual–textual cues and coupled degradations of inscriptions; (2) correcting the recognized script with a specialized CLM yields a clear gain in 1-NED, highlighting the importance of linguistic priors for textual authenticity; (3) adding RAG achieves the best performance, indicating that corpus-level retrieval further improves robustness and generalization.

\begin{table}[t]
\centering
   \caption{Ablation study of different planning strategies for inscription restoration. ``Random" refers to random tool invocation from the toolkit. ``Fixed" denotes predefined Scheme A (Denoising-Completion) and Scheme B (Denoising-Completion-Imitation). ``Experience-guided" customizes the restoration scheme based on distilled experience priors. The best and the second best results are \textbf{highlighted} and \underline{underlined}.}
\setlength{\tabcolsep}{2.8mm} %2.2mm
\renewcommand{\arraystretch}{1}
\adjustbox{width=0.9\linewidth}{
   \begin{tabular}{ccccc}
       \toprule
                 \multirow[c]{2}{*}{\vspace*{-1.5mm}\centering Strategy / Metric} & \multicolumn{3}{c}{Quality} & End-to-End \\
\cmidrule[0.5pt](lr){2-4} \cmidrule[0.5pt](lr){5-5}
                  & PSNR $\uparrow$ & SSIM $\uparrow$ & LPIPS $\downarrow$ & 1-NED $\uparrow$\\
        
%%%%%%%%%%%%%% 
\midrule[0.5pt]      
Random & 18.53 & 0.9078 &0.0869 & 0.7702 \\
Fixed (Scheme A)         &\underline{21.19} &\underline{0.9605} &\underline{0.0371} &0.8814\\
Fixed (Scheme B)          &20.78 &0.9526 &0.0401 &\underline{0.8935} \\
\cellcolor{blue!8}Experience-guided (Ours)        &\cellcolor{blue!8}\textbf{22.14} &\cellcolor{blue!8}\textbf{0.9684} &\cellcolor{blue!8}\textbf{0.0254} &\cellcolor{blue!8}\textbf{0.9069} \\

 \bottomrule
       \end{tabular}
   }
\label{table:plan}
\end{table}

\begin{table*}[t]
\centering
   \caption{Ablation studies of multi-perspective evaluation module. The best and the second best results are \textbf{highlighted} and \underline{underlined}.}
\setlength{\tabcolsep}{2.8mm} %2.2mm
\renewcommand{\arraystretch}{1}
\adjustbox{width=1\linewidth}{
   \begin{tabular}{ccc ccc cccc c}
       \toprule
                 \multicolumn{2}{c}{Automatic Metric}
                 & \multicolumn{1}{c}{Human Feedback}
                 & \multicolumn{7}{c}{Quality}
                 & \multicolumn{1}{c}{End-to-End} \\
\cmidrule[0.5pt](lr){1-2}\cmidrule[0.5pt](lr){3-3} \cmidrule[0.5pt](lr){4-10}\cmidrule[0.5pt](lr){11-11}
                 Text Authenticity & Style Consistency  & Score of Human Experts & PSNR $\uparrow$ &SSIM $\uparrow$ & LPIPS $\downarrow$ & CLIP-IQA $\uparrow$ & MUSIQ $\uparrow$ & MANIQA $\uparrow$ & NIMA $\uparrow$ & 1-NED $\uparrow$  \\
                 \cmidrule[0.5pt](lr){1-3}\cmidrule[0.5pt](lr){4-11}
%%%%%%%%%%%%%% 
% \midrule[0.5pt]      
\ding{55} & \ding{55}  & \ding{55}              &21.48 &0.9593 &0.0305 &0.8976 &53.63 &0.4454 &0.5549 &0.8969 \\
\ding{51} & \ding{55} & \ding{55}          &21.59 &0.9616 &0.0292 &0.8974 &53.66 &0.4465 &0.5552 &0.9026\\
\ding{55} & \ding{51}  & \ding{55}         &21.73 &0.9632 &0.0284 &0.8979 &53.70 &0.4477 &0.5559 &0.8994\\
\ding{51} & \ding{51}  &  \ding{55}         &\underline{22.02} &\underline{0.9668} &\underline{0.0279} &\underline{0.8983} &\underline{53.73} &\underline{0.4492} &\underline{0.5566} &\underline{0.9041} \\
\cellcolor{blue!8}\ding{51} &\cellcolor{blue!8}\ding{51}  &\cellcolor{blue!8}\ding{51}     &\cellcolor{blue!8}\textbf{22.14} &\cellcolor{blue!8}\textbf{0.9684} &\cellcolor{blue!8}\textbf{0.0254} &\cellcolor{blue!8}\textbf{0.9004} &\cellcolor{blue!8}\textbf{53.98} &\cellcolor{blue!8}\textbf{0.4553} &\cellcolor{blue!8}\textbf{0.5576} &\cellcolor{blue!8}\textbf{0.9069} \\
 \bottomrule
       \end{tabular}
   }
\label{table:metric}
\end{table*}

\subsubsection{The Effect of Adaptive Planning}

We next evaluate how adaptive planning in the \textbf{Conceive} stage affects restoration quality. As illustrated in Fig.~\ref{fig:sequence}, improper tool ordering can severely degrade results, e.g., skipping background denoising and directly applying stroke completion compromises glyph integrity and style consistency (Fig.~\ref{fig:sequence}(a)), while inappropriate use of completion and imitation leads to over-restoration with fake structures (Fig.~\ref{fig:sequence}(b)). These cases highlight that effective planning is non-trivial and must be guided by domain knowledge and accumulated restoration experience.
To quantify this effect, we compare three planning strategies on Testing Set S: ``Random", ``Fixed", and our experience-guided adaptive planning based on distilled priors. As reported in Tab.~\ref{table:plan}, the experience-guided strategy achieves superior scores across all three image-quality metrics and spotting accuracy, confirming the benefit of adaptive planning for handling dual-modal cues and complex degradations in inscriptions.

\begin{figure}[t]
\centering         
\includegraphics[width=1.0\linewidth]{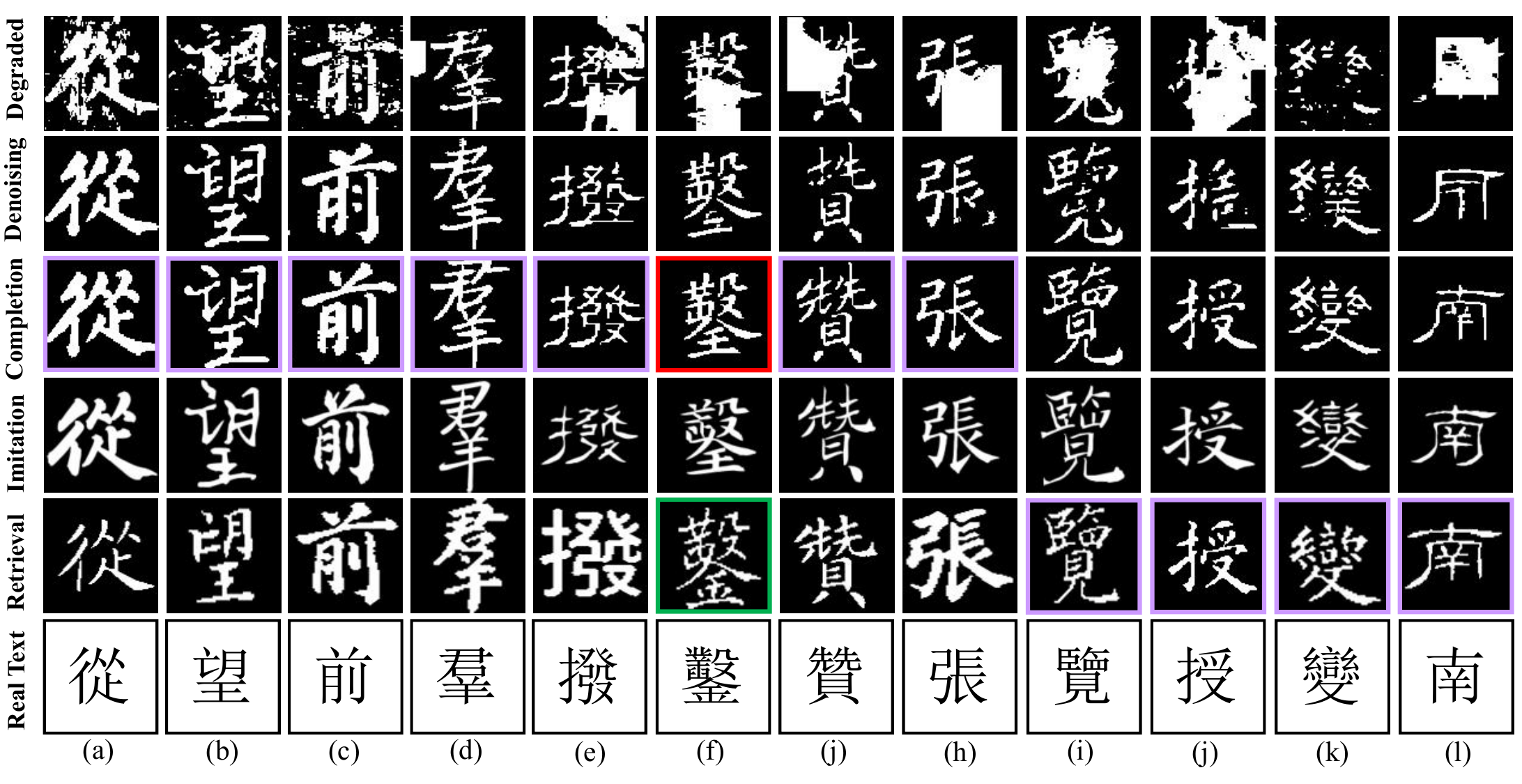}
\caption{Comparison of restoration outcomes across different tools. (a)-(d), (e)-(h), (i)-(l) correspond slightly, middely, and severely degraded character. \textcolor{red}{Red} and \textcolor{darkgreen}{green} borders indicate the restorations preferred by EpiAgent and human experts, respectively, while \textcolor{lightpurple}{purple} borders indicate cases where their choices coincide.}
\label{fig:char}
\end{figure}

\subsubsection{The Impact of Specialized Restoration Toolkit}

Fig.~\ref{fig:char} provides a character-level comparison of restorations obtained with different tools in the specialized toolkit. In Fig.~\ref{fig:char}(a)–(h), the background denoising and stroke completion tools effectively handle slight to medium degradation. Specifically, denoising removes scattered noise and artifacts, while the completion tool then refines and reconstructs damaged stroke structures. However, under severe spalling or complete absence, as in Fig.~\ref{fig:char}(i)–(l), denoising and completion alone become insufficient since the limited local evidence makes inpainting ill-posed. In such cases, font imitation and character retrieval provide reliable alternatives. 
Nonetheless, these two tools can still produce undesirable results, as shown in Fig.~\ref{fig:char}(h), highlighting the necessity of dynamic tool combination.

\subsubsection{The Role of Evaluation and Refinement}

We investigate the multi-perspective evaluation used in the \textbf{Reevaluate} stage. From Tab.~\ref{table:metric}, we obtain three observations: (1) adding the text authenticity metric markedly improves end-to-end spotting accuracy, indicating that target prediction deviations provide an effective signal to refine each restoration pass; (2) incorporating the style consistency metric further boosts visual quality across all seven full- and no-reference IQA measures, showing that explicitly scoring calligraphic coherence helps the agent better preserve glyph aesthetics; (3) expert review yields the highest gains, as expert-in-the-loop feedback more faithfully reflects actual restoration quality and injects valuable domain knowledge.

\begin{figure}[!t]
\centering         
\includegraphics[width=1.0\linewidth]{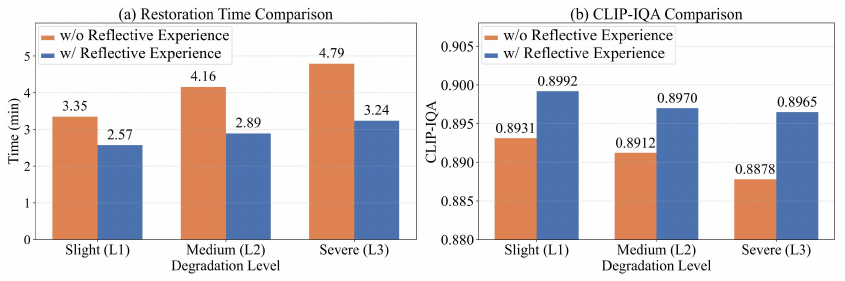}
\caption{Quantitative comparison of restoration time and CLIP-IQA with and without reflective experience.}
\label{fig:reflection}
\end{figure}

Fig.~\ref{fig:reflection} examines the effect of disabling the refinement mechanism, i.e., evaluating restorations without recording and reusing feedback. Removing this refinement mechanism leads to a clear increase in average restoration time and a decline in quality, confirming that accumulated reflective experience is both reusable and beneficial for enhancing the inscription restoration capability of EpiAgent.

\section{Conclusion}

We introduced \textbf{EpiAgent}, an agent-centric system for ancient inscription restoration that operationalizes the workflow of epigraphers. An LLM-based central planner coordinates the dynamic four-stage process, enabling hierarchical closed-loop decision-making that safeguard textual authenticity and visual fidelity. Experiments and ablation studies show that EpiAgent consistently surpasses strong baselines, highlighting the effectiveness of our domain-aware design. Beyond inscriptions, our framework offers a concrete blueprint for integrating agentic AI into cultural heritage preservation.

{
    \small
    \bibliographystyle{ieeenat_fullname}
    \bibliography{main}
}


\end{document}